\documentclass[journal]{IEEEtran}

\ifCLASSINFOpdf
\else
\fi
\usepackage{amsmath}
\usepackage{cite}
\usepackage{graphicx}
\graphicspath{ {Images/} }
\usepackage{subfigure}
\usepackage{array,multirow}
\usepackage{verbatim}
\usepackage{mathtools}
\usepackage{caption}
\newcolumntype{C}[1]{>{\centering\let\newline\\\arraybackslash\hspace{0pt}}m{#1}}
\newcolumntype{T}[1]{>{\let\newline\\\arraybackslash\hspace{0pt}}m{#1}}
\newcommand{\etal}{\textit{et al}.}
\usepackage{stackengine}
\newcommand\barbelow[1]{\stackunder[1.2pt]{$#1$}{\rule{.8ex}{.075ex}}}

\usepackage{booktabs}

\begin{document}
%
\title{SISC: End-to-end Interpretable Discovery Radiomics-Driven Lung Cancer Prediction via Stacked Interpretable Sequencing Cells}
%
%
%

\author{Vignesh~Sankar*,~\IEEEmembership{Student Member,~IEEE,}
        Devinder~Kumar*,~\IEEEmembership{Student Member,~IEEE,}
        David~A.~Clausi,~\IEEEmembership{Senior Member,~IEEE,}
        Graham W.~Taylor,~\IEEEmembership{Member,~IEEE,}
        and~Alexander~Wong,~\IEEEmembership{Senior Member,~IEEE,}
\thanks{*Equal Contribution; Vignesh Sankar, Devinder Kumar, David Clausi and Alexander Wong are with the Department
of Systems Design Engineering, University of Waterloo, Waterloo,
ON, Canada N2L3G1 e-mail: (vignesh.sankar,d22kumar,dclausi,a28wong)@uwaterloo.ca.}
\thanks{Graham W. Taylor is with the School of Engineering, University of Guelph, Guelph, ON, Canada and the Vector Institute, Toronto, Canada.}
}

%
%

\markboth{IEEE Transactions of Biomedical Engineering}%
{Sankar \MakeLowercase{\textit{et al.}}:}
%



\maketitle

\begin{abstract}

Objective: Lung cancer is the leading cause of cancer-related death worldwide. Computer-aided diagnosis (CAD) systems have shown significant promise in recent years for facilitating the effective detection and classification of abnormal lung nodules in computed tomography (CT) scans. While hand-engineered radiomic features have been traditionally used for lung cancer prediction, there have been significant recent successes achieving state-of-the-art results in the area of discovery radiomics. Here, radiomic sequencers comprising of highly discriminative radiomic features are discovered directly from archival medical data.  However, the interpretation of predictions made using such radiomic sequencers remains a challenge. Method: A novel end-to-end interpretable discovery radiomics-driven lung cancer prediction pipeline has been designed, build, and tested. The radiomic sequencer being discovered possesses a deep architecture comprised of stacked interpretable sequencing cells (SISC). Results: The SISC architecture is shown to outperform previous approaches while providing more insight in to its decision making process. Conclusion: The SISC radiomic sequencer is able to achieve state-of-the-art results in lung cancer prediction, and also offers prediction interpretability in the form of critical response maps. Significance: The critical response maps are useful for not only validating the predictions of the proposed SISC radiomic sequencer, but also provide improved radiologist-machine collaboration for effective diagnosis. 

\end{abstract}

\begin{IEEEkeywords}
lung, interpretable, nodule, radiomics, cancer, discovery radiomics
\end{IEEEkeywords}

%
\IEEEpeerreviewmaketitle

\section{Introduction}
%

Lung cancer is the most diagnosed form of cancer and the leading cause of cancer-related death worldwide~\cite{nagai2017cancer}.  Approximately 225,000 new cases in the United States appear each year and it leads to \$12 billion in annual health care costs~\cite{LungCancerStat}. In fact, the number of individuals suffering from lung cancer is greater than all individuals suffering from breast, colon, and prostate cancer combined.

Early detection and diagnosis of lung cancer is critical as it can significantly reduce mortality rates.  In particular, low dose computed tomography (CT) imaging has proven to be one of the most effective ways of detecting early stages of lung cancer~\cite{zanon2017early}. However, one of the key challenges with lung cancer detection and diagnosis using CT imaging is that the manual screening of CT scans is a very laborious and time-consuming process. This is true even for expert radiologists, given the large amount of imaging data that needs to be analyzed. As such, there is a growing need for computer-aided diagnosis (CAD) systems which can assist radiologists in cancer detection in a fast yet accurate manner. Such systems can provide automatic identification of the cancerous nodules in CT scans along with useful information about their malignancy characteristics.

\begin{figure*}[h]
\centering
\includegraphics[scale=0.5]{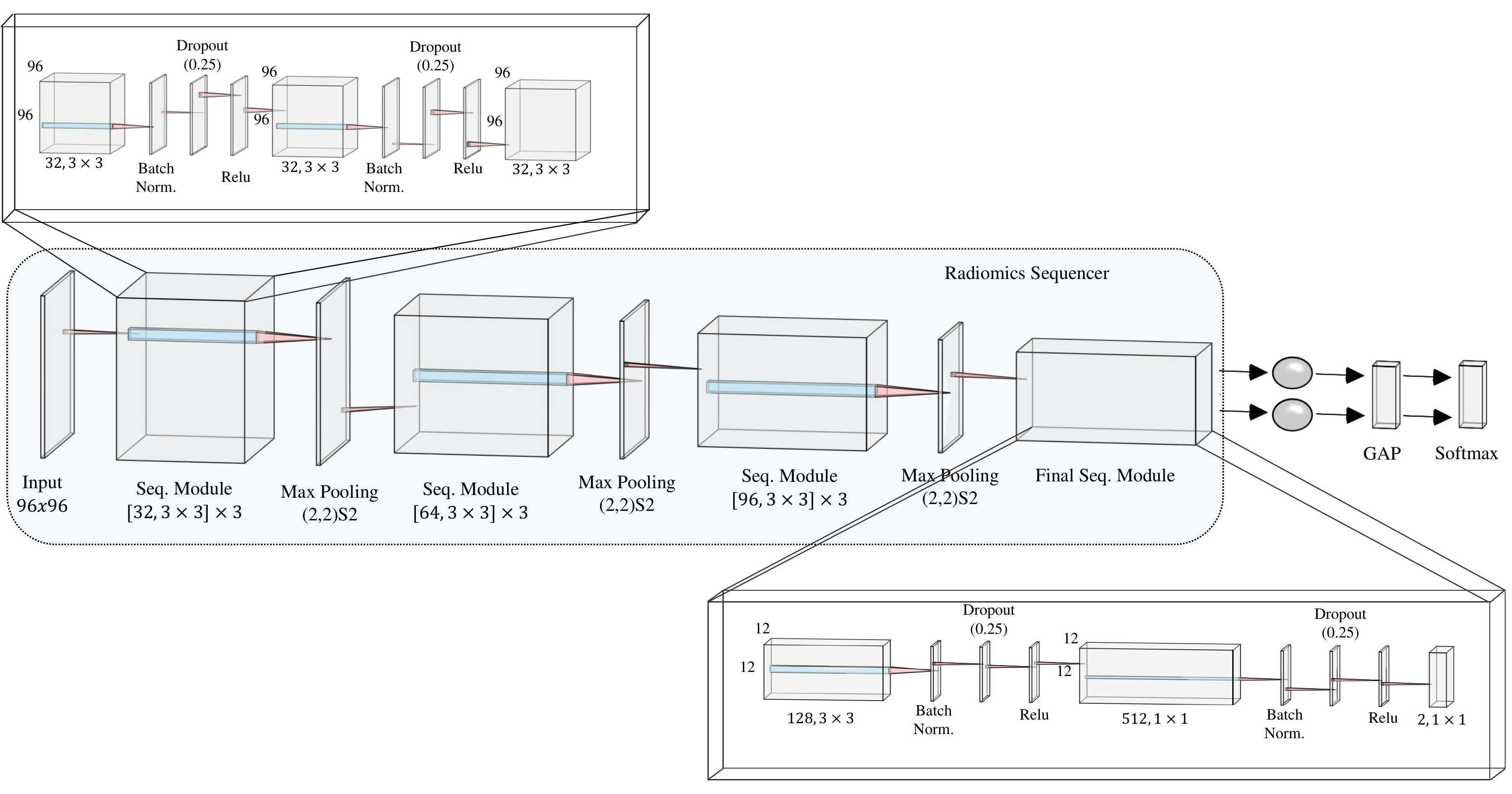}
\caption{Overview of the proposed deep stacked interpretable sequencing cell (SISC) architecture used as the radiomic sequencer within a discovery radiomics framework. The SISC radiomic sequencer is formed by stacking interpretable sequencing cells together, each comprised of different specialized convolutional layers along with max-pooling and dropout operations. A typical interpretable sequencing cell consists of a block of convolutional layers followed by batchnorm, dropout, and ReLU layers, repeated three times. The final interpretable sequencing cell is flexible and can be changed based on the input and task.  Here, the final layer in the last cell consists of two 1$\times$1 convolutions as this study is focused on binary lung cancer classification.}
\label{ISCfig}
\end{figure*}

Until recently, state-of-the-art CAD systems relied on radiomics-based approaches built from hand-engineered radiomic features. These features are designed to characterize cancer phenotypes in a high-dimensional feature space that facilitates tumor discrimination.  
The extracted radiomic sequence consists of a large number of predefined, hand-engineered features for capturing image-based traits such as intensity, texture, and shape. These hand-engineered features can greatly limit the ability of the radiomic sequence to fully characterize the unique traits of different forms of cancer. More recently, the concept of discovery radiomics~\cite{chung2015discovery,shafiee2015discovery,kumar2017discovery,wong2015discovery,karimi2017discovery} was shown to achieve state-of-the-art performance. In discovery radiomics, the radiomic sequencers generate highly-discriminative quantitative radiomic features which are discovered, rather than hand-engineered, from vast amounts of available archival medical data. 
In particular, radiomic sequencers with deep convolutional architectures that are discovered in an end-to-end manner were shown to consistently achieve state-of-the art performance in medical imaging analysis. The use of such radiomic sequencers within the discovery radiomics framework is particularly effective for lung cancer prediction using CT imaging. This is due to the availability of very large annotated CT scan data sets such as LIDC-IDRI~\cite{armato2011lung}, which enables highly discriminative radiomic features to be discovered directly from this wealth of data. 

Despite the effectiveness of discovery radiomics-based approaches from a diagnostic performance perspective, a key challenge that still remains is the difficulty in interpreting the rationale behind their predictions. As such, one can view such radiomics-based approaches as 'black box', and the lack of transparency in their decision-making processes makes it difficult for radiologists to verify, validate, and ultimately trust the predictions being made. To enable the widespread adoption of discovery radiomics within CAD systems, one needs to improve radiologists' trust by providing interpretable reasoning behind the predictions made by radiomic sequencers. 

Motivated by this, we propose an end-to-end interpretable discovery radiomics-driven lung cancer prediction pipeline.  Specifically, the main contributions of our approach are:
\begin{itemize}
    \item the introduction of SISC architecture (Fig.~\ref{ISCfig}), comprising of interpretable sequencing cells, for building radiomic sequencers with state-of-the-art performance for lung cancer prediction, and 
    \item interpretable lung cancer predictions in the form of critical response maps generated through a stack of interpretable sequencing cells which highlight the key critical regions leveraged in the prediction process (Fig.~\ref{overview}).
\end{itemize}


\section{Background}
\label{background}
There has been significant interest and a wealth of literature in the past decade, especially in lung cancer detection and diagnosis~\cite{xie2018fusing,de2018classification,zhu2018deeplung,shen2018interpretable,liu2018multi,wang2018improved}. CAD systems designed for automated lung cancer diagnosis can be divided into two main stages: 1) Lung nodule detection and segmentation; and 2) malignancy classification of segmented nodules.

The first stage consists of processing CT images to isolate the lung region~\cite{shen2015automated} and search for potential pulmonary nodule candidates~\cite{suzuki2009supervised,duggan2015technique}. The process involves finding the location of the pulmonary nodules in a given CT scan slice by defining its position and contour~\cite{ronneberger2015u}. The candidate detection step is usually followed by false positive reduction algorithms~\cite{suzuki2003massive,suzuki2005false,suzuki2003effect} to control the rate of candidate generation.

The second stage is the nodule classification process, where the detected nodules are classified as either malignant or benign.  In particular, there has been significant interest in the concept of radiomics, which involves the high-throughput extraction and analysis of a large amount of quantitative features from medical imaging data to characterize tumor phenotypes in a quantitative manner.  Different radiomic feature extraction techniques and algorithms have been proposed in the literature to perform the classification task, and can be split into two broad categories: 1) hand-engineered radiomic features, and 2) learned/discovered radiomic features.  

\subsection{Hand-Engineered Radiomic Features}
Traditional radiomics-based approaches leveraged pre-defined, hand-engineered features designed with the help of radiologists  ~\cite{kaya2015weighted,buty2016characterization,huang2017added}. Such hand-engineered features typically capture generic image-based traits such as intensity, texture, and shape. For example, Way \etal~\cite{way2006computer} used the segmented nodules to extract texture features and classified them using linear discriminant classifier. El-Baz \etal~\cite{el20113d} used the shape of nodules as features whereas, Han \etal~\cite{han2013texture} used 3D texture analysis to extract discriminative features for nodule classification. Dhara \etal~\cite{dhara2016combination} used an ensemble of margin-based, shape-based and texture-based features extracted from the segmented nodule to form the feature set. Support vector machines were then used to classify the nodule as malignant or benign using this radiomic feature set. 

Aydin \etal~\cite{kaya2015weighted} used radiologist-provided evaluations to develop a weighted rule-based algorithm using an ensemble of classifiers to predict malignancy score. Recently, Robherson \etal~\cite{de2018classification} used taxonomic diversity index and mean plylogenetic distance to calculate the malignancy score. In general, the most commonly used feature extraction methods are histogram of oriented gradients (HOG)~\cite{dalal2005histograms,firmino2016computer} and local binary patterns (LBP)~\cite{ojala2002multiresolution}. Different machine learning classifiers such as support vector machines (SVM)~\cite{han2015texture}, random forests~\cite{armato2016lungx} and weighted nearest neighbour (NN)~\cite{reeves2016automated} were used with the hand-crafted features for classification. One limitation to hand-engineered radiomic features is that they are typically designed based on generic image-based features that are not customized for the task of lung cancer detection and diagnosis, and thus can greatly limit the classifier's ability to fully characterize the unique traits of different forms of cancer.

\subsection{Learned/Discovered Radiomic Features}
With the recent advances and success in machine learning, particularly deep learning~\cite{krizhevsky2012imagenet}, many researchers have started to leverage such approaches for medical image analysis~\cite{zhao2017hybrid,shen2015multi,li2016pulmonary,setio2016pulmonary,chung2015discovery}.  This has led to significant interest in the concept of discovery radiomics, where high-dimensional quantitative radiomic features are learned and discovered directly from the wealth of medical imaging data available.  Such discovered radiomic sequences enable a much more customized characterization of tumor phenotype.  As one of the first deep learning-driven discovery radiomics approaches for the purpose of lung cancer classification, Kumar \etal~\cite{kumar2015lung} learned deep radiomic features using an auto-encoder radiomic sequencer for the purpose of classifying lung nodules as malignant or benign. Currently, deep convolutional radiomic sequencers have shown tremendous success for the task of nodule classification. With sufficient architectural depth, these radiomic sequencers enable a robust classification framework that can accommodate the variability found in characteristics of the lung nodules. For example, Mario \etal~\cite{buty2016characterization} combined shape and appearance radiomic features with learnt radiomic features from a deep convolutional radiomic sequencer to form a combined radiomic sequence that is then used to feed a random forest classifier.  Shen \etal~\cite{shen2017multi} proposed a multi-scale convolutional radiomic sequencer to obtain radiomic features for the purpose of nodule classification. Zhao \etal~\cite{zhao2017hybrid} proposed a hybrid radiomic sequencer to combine deep convolutional features with LBP and HOG features. A novel multi--scale, multi--view radiomic sequencer architecture was proposed by Liu \etal~\cite{liu2018multi} for nodule classification. The 3D nature of the CT scans has also inspired the use of 3D convolutional radiomic sequencer architectures. Liu \etal~\cite{liu2017pulmonary} proposed a novel 3D convolutional architecture for lung nodule classification. Zhu \etal~\cite{zhu2017deeplung} proposed an end--to--end automated lung cancer detection framework called ”DeepLung”, where 3D dual path net feature extraction was introduced. Dey \etal~\cite{dey2018diagnostic} proposed a two-pathway 3D convolutional architecture to capture both local and global characteristics. 

\begin{figure*}[h]
\centering
\includegraphics[scale=0.52]{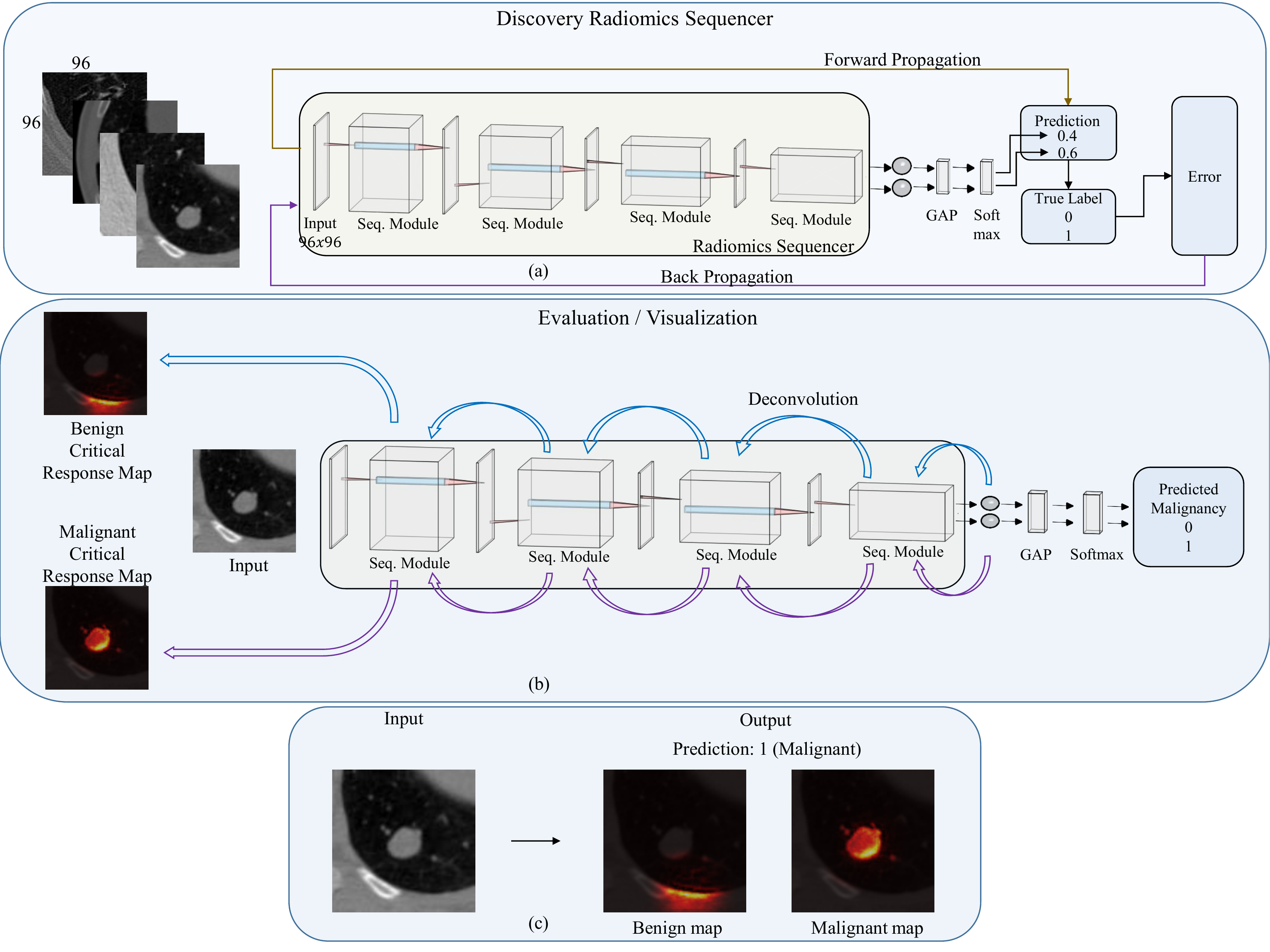}
\caption{Overview of the end-to-end interpretable discovery radiomics-driven framework for lung cancer prediction. Part (a) shows the sequencer discovery process, where a specialized radiomic sequencer, comprised of a deep stack of interpretable sequencer cells, is discovered for the given set of CT lung nodule data.  Part (b) presents the cancer prediction process, where the discovered radiomic sequencer is used to make a prediction based on CT data and how interpretable critical response maps are generated through the stack of sequencer cells. In part (b), the input CT data of a new patient is fed into the radiomic sequencer to generate a radiomic sequence and perform prediction on whether it is a benign and malignant case.  To generate the critical response map, the output of the last layer in the sequencing cell of the radiomic sequencer is backpropagated through each sequencing cell using the method described in Section~\ref{visualization} for each of the possible prediction states (benign and malignant). As such, we obtained two critical response maps, each highlighting the critical regions used by the sequencer for making predictions regarding whether the given input nodule is benign or malignant. The last part (c) is the interface seen by the end user, which shows the given input, the prediction and evidence been used to obtain the particular prediction through critical response map.}
\label{overview}
\end{figure*}

\subsection{Interpretability}
\label{litr_Interpretability}
While highly complex deep convolutional radiomic sequencer architectures can achieve state-of-the-art performance, one of the biggest limitations to leveraging such sequencers is that they are very difficult to interpret.  As such, researchers have in recent years explored different approaches for better understanding the inner workings of such architectures. The current approaches can be grouped into two categories. The first group of approaches attempts to understand the global decision making process of the deep convolutional architectures by identifying inputs which maximize the outputs in the architecture~\cite{erhan2010understanding,baehrens2010explain,goodfellow2009measuring,yosinski2015understanding}. The second group of approaches provides an explanation for the prediction being made by generating attentive maps for the input image. The attentive maps highlight the attentive regions used by the architecture for making a particular prediction.  Since it is important for clinicians interacting with a CAD system to be able to visualize critical regions linked to cancer and for the purpose of justifying the computer-aided lung cancer prediction, we will discuss the second group of approaches in detail below. 

Simonyan \etal~\cite{simonyan2013deep} used derivatives of the class score found by back-propagation for each pixel to create the class saliency map. Zeiler \etal~\cite{zeiler2014visualizing} and Springenberg \etal~\cite{springenberg2014striving} used a deconvolution-based method and gradient-based method, respectively, to project activations back to the input space. A limitation of both methods is that there is no class information to their visualizations. Zhou \etal~\cite{zhou2016learning} proposed to use global average pooling layers in CNN to create what they refer to as Class Activation Maps (CAM). Zintgraf \etal~\cite{zintgraf2017visualizing} proposed to use multi-variate conditional sampling to visualize the predictions made by deep convolutional architectures and highlight the input image's pixel which contributes for and against a particular class. Kumar \etal~\cite{kumar2017explaining} proposed the concept of CLEAR (CLass-Enhanced Attentive Response) maps which can provide the per-class attentive interest levels on each input image's pixel in the given deep convolutional architecture.

\begin{figure*}[h]
\includegraphics[scale=0.5]{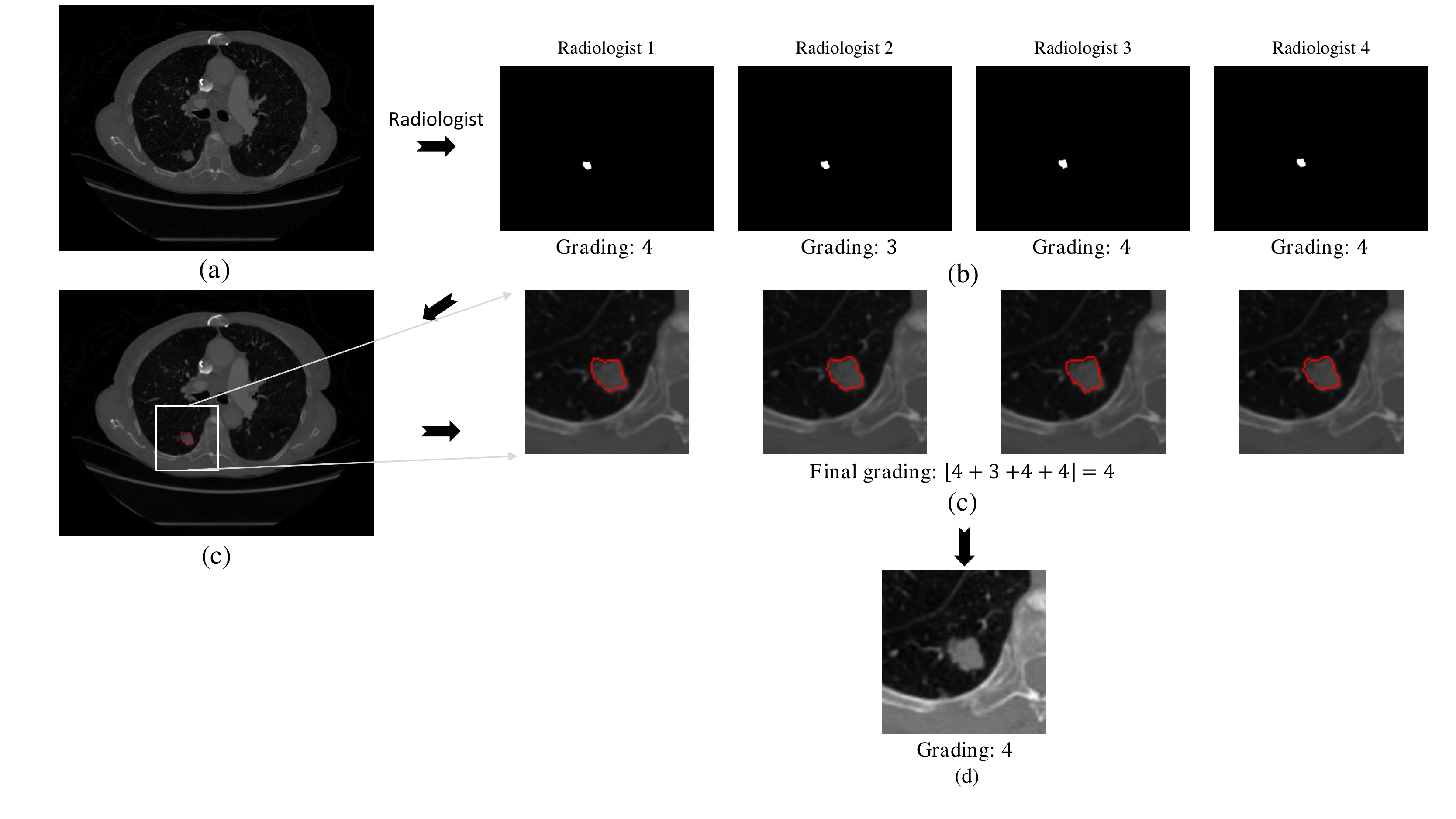}
\centering
\caption{Data preparation flow for i) computing malignancy score, and ii) obtaining an 96$\times$96 nodule image directly from the raw LIDC-IDRI dataset. For a given lung CT slice image (a), the LIDC-IDRI dataset contains lung nodule masks from up to four radiologist (b). We use these masks to locate the contour of the nodule in the CT slice image.  As can be seen in the zoomed-in images in (c), the contours of the mask differ due to the inter-observer variability. To mitigate this, we selected an sub-image of 96$\times$96 from the average center pixel obtained from the radiologist masks to create a single 96$\times$96 lung nodule image (d). For computing the malignancy score (as shown in (d)), we average the scores given by the radiologists as indicated in part (c). A detailed discussion on data preparation is discussed in Section III part A.}
\label{dataprep}
\end{figure*}

\section{Methodology}
\label{methodology}
This section describes the design and implementation of a radiomics sequencer. The modular design of the SISC architecture (see Section~\ref{isc}) enables a significant reduction in the design search space while improving classification performance.  Furthermore, the introduction of interpretable sequencing cells allow us to achieve end-to-end interpretability through the generation of critical region maps to aid the clinician in the decision-making process (see Section~\ref{visualization}).  The LIDC-IDRI dataset (see Section~\ref{dataset1}) is used to validate the proposed radiomic sequencer architecture.

\subsection{Dataset}
\label{dataset1}
The Lung Image Database Consortium (LIDC) and Image Database Resource Initiative (IDRI)~\cite{armato2011lung} published a structured and categorized repository of computed tomography (CT) scans to assist the development of CAD methods for automated lung cancer diagnosis. The dataset consists of 1018 thoracic CT scans, where each scan is processed by four radiologists at both blinded and un--blinded stages. In the blinded stage, each radiologist reviews the CT scans without inputs from other radiologists. In the second, un--blinded stage, each radiologist is shown the results of other radiologists from the blinded stage and is given a chance to change their initial evaluations. The two stage process was designed to provide the best estimate of the nodule characteristics.

The suspected lung lesions in the LIDC dataset are divided into three categories: i) Non-nodule$\geq$3mm, ii) Nodule$\geq$3mm, and iii) Nodule$<$3mm, where the diameter is measured as the length of the lesion's longest axis. For each category, different nodule characteristics are included in the dataset. Similar to previous methods, we decided to only use Nodule$\geq$3mm. For the Nodule$\geq$3mm category, the required malignancy score along with the nodule location and contour information by at least one radiologist are included. Therefore, the Nodule$\geq$3mm category is used for our experiments. A total of $2669$ nodules are reported in the dataset under the Nodule$\geq$3mm category.

For each nodule, its characteristics are provided by at most four radiologists. The final malignancy score for each nodule is obtained by combining the scores from all of the radiologists. As suggested in~\cite{han2013texture}, the average score rounded to the nearest integer was taken as the final malignancy score. In each slice of the given nodule, a $96\times96$ window was cropped at the nodule center. The size was determined to accommodate for all the variability in the nodule contour and also to include sufficient background information. The same malignancy score was assigned for all the slices in the given nodule. A total of $14,433$ nodule images along with their corresponding malignancy score were extracted from the dataset.

\subsection{Interpretable Sequencing Cells}
\label{isc}

A modular design strategy was leveraged to construct the proposed SISC radiomics sequencer, where the underlying architecture is comprised of a deep stack of interpretable sequencing cells with similar micro-architectures.  More specifically, an intepretable sequencing cell as introduced in this study comprises of a block of convolutional layers along with max-pooling and dropout operations, all optimized using the available data. The proposed SISC radiomic sequencer is then constructed by stacking the interpretable sequencing cells together in a depth-wise manner. The aim is to reduce the design search space while improving classification accuracy, thus enabling optimized design of sequencer architectures in a more predictable manner. The micro-architecture of an interpretable sequencing cell is defined by three convolutional layers separated by batch normalization~\cite{ioffe2015batch} and drop--out~\cite{srivastava2014dropout}. Furthermore, the ReLU~\cite{glorot2011deep} activation is used after each convolutional layer. The interpretable sequencing cell is optimized by sharing the same architectural values. For example, all the dropout layers in a given interpretable sequencing cell have the same dropout rate.

\begin{figure*}[h]
\includegraphics[scale=0.5]{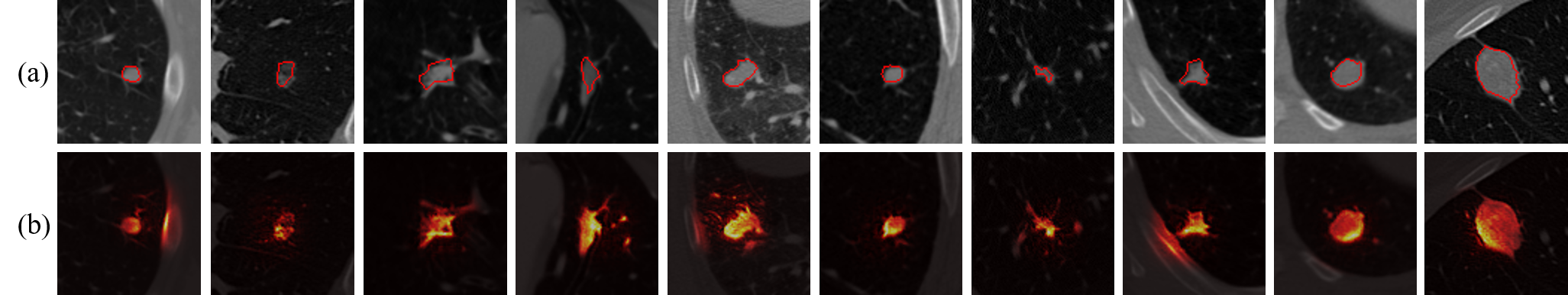}
\centering
\caption{Example critical response maps for malignant cases. (a) original 96$\times$96 malignant nodule sub-image taken from lung CT slices with the radiologist-provided best contours for a given patient CT slice image, and (b) corresponding critical response maps showing the malignant critical regions been used for correctly predicting malignant nodules. It can be seen that for almost all the example cases, the proposed SISC radiomic sequencer uses clinically relevant markers when achieving correct predictions.  Therefore, the use of critical response maps can potentially improve the overall confidence of the clinician on the discovered SISC radiomic sequencer.}
\label{Deconv1}
\end{figure*}

\begin{figure*}[h]
\includegraphics[scale=0.5]{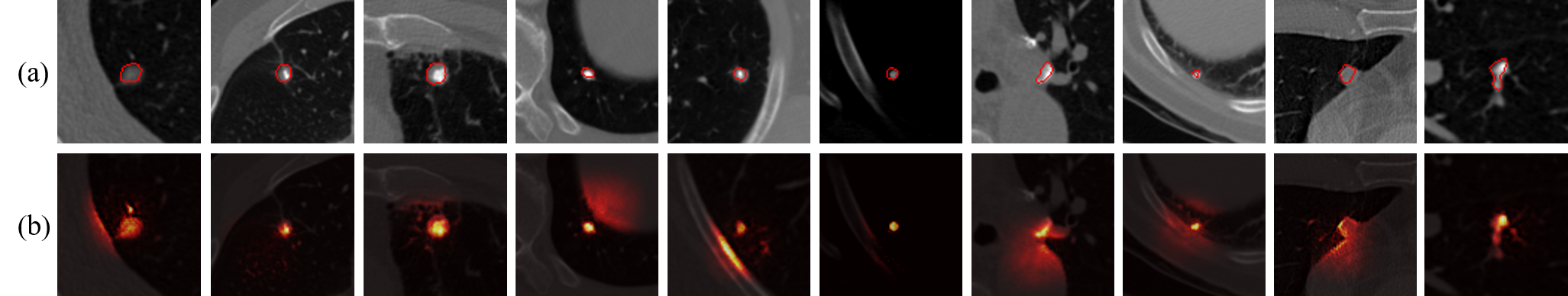}
\centering
\caption{
Example critical response maps for benign cases. (a) original 96$\times$96 malignant nodule sub-image taken from lung CT slices with the radiologist-provided best contours for a given patient CT slice image, and (b) corresponding critical response maps showing the benign critical regions been used for correctly predicting benign nodules. It can be seen that for almost all the example cases, the proposed SISC radiomic sequencer uses clinically relevant markers when achieving correct predictions.  Therefore, as with the previous figure, the use of critical response maps can potentially improve the overall confidence of the clinician on the discovered SISC radiomic sequencer.}
\label{Deconv0}
\end{figure*}

In this study, the proposed SISC radiomic sequencer is comprised of four interpretable sequencing cells stacked together in a depth-wise manner as shown in Fig~\ref{ISCfig}. The number of channels is increased as we go deeper into the SISC architecture whereas the size of the kernel is fixed for the first three cells. Each cell is followed by a max pooling layer. The dropout, batch normalization parameters and number of cells are optimized with the LIDC-IDRC dataset. The final cell of the SISC radiomic sequencer is defined as shown in Fig.~\ref{ISCfig}. The number of kernels in the final convolutional layer of the final cell is equivalent to the total number of classes to enable end-to-end interpretability via critical response maps, which will be further discussed in the following section. The final convolutional layer is followed by a global average pooling layer, which is then followed by a softmax output layer.

The proposed SISC radiomic sequencer formed by the stacking of interpretable sequencing cells with learned parameters is shown in Fig~\ref{ISCfig}. We can observe that the repeated modular approach allows us to compactly define the sequencer with a minimum number of configurable architectural parameters. The modular approach also leads to state-of-the-art results as described in Section~\ref{experimentsandresults}.

\subsection{Interpretability through Critical Response Maps}
\label{visualization} 

To enable interpretability and explainability in the decision-making process of the proposed SISC radiomic sequencer, we take inspiration from~\cite{kumar2017explaining} and~\cite{kumar2017opening} and introduce an approach where critical response maps are generated through the entire stack of interpretable sequencing cells. An critical response map provides spatial insights on critical regions in the CT scan and their level of contribution to a particular prediction made.  Here, an individual critical response map is generated for each possible prediction (benign and malignant).

Using these critical response maps, the clinician can not only validate the the evidence behind the predictions made using the proposed SISC radiomic sequencer, but the maps also help in locating relevant regions in the CT scan responsible for either a malignant or benign nodule prediction. An example pair of critical response maps can be seen in Fig.~\ref{overview}. 


For example, in the case of a malignant nodule, a successful and reasonable prediction should lead to the malignant critical map highlighting the nodule regions.  As such, critical response maps may potentially help radiologists to have greater confidence in the CAD system and in aiding them with their clinical diagnosis decisions.

The critical response map generation process can be described as follows. Let the critical response maps $A(x|c)$ for a given CT scan slice image $x$ for each prediction $c$ be computed via back-propagation from the last layer of the last interpretable sequencing cell in the proposed SISC radiomic sequencer. The notation used in this study are based on the study done by Kumar et. al.\cite{kumar2017explaining} for consistency. 
As shown in Fig~\ref{ISCfig}, the last layer in the interpretable sequencing cell at the end of the proposed SISC radiomic sequencer contains $N=2$ nodes, equal to the number of possible predictions (i.e., benign and malignant). The output activations of this layer are followed by global average pooling and then a softmax output layer. So, to create the critical maps for each possible prediction, the back-propagation starts with the individual prediction nodes in the last layer to the input space. For a single layer $l$, The deconvolved output response $\hat{r}_l$ is given by,

\begin{equation}
	\hat{h}_l = \sum_{k=1}^{K} f_{k,l}*p_{k,l}
\end{equation}
\noindent where $f_k$ is the feature map and $p_l$ is the kernel of layer $l$. the symbol $*$ represents the convolution operator. For simplicity, The convolution and summation can be combined as $\hat{h}_l = D_lf_l$. Therefore, the critical response map $A(\barbelow{x}|c)$, for a given prediction $c$ is defined as,

\begin{equation}
A(\barbelow{x}|c) = D_1M'_1D_2M'_2....D_{L-1}M'_{L-1}D^{c}_LF_L.
\end{equation}
Where $M'$ is the un-pooling operation as described in~\cite{zeiler2010deconvolutional} and $D^{c}_L$ is the convolution operation at the last layer with kernel $p_L$ replaced by zero except at the $c^{th}$ location corresponding to the prediction $c$.


\section{Experiments and Results}
\label{experimentsandresults}

In this section, we will evaluate and discuss the efficacy for the proposed SICS radiomic sequencer for the purpose of lung cancer prediction on two main fronts: i) cancer prediction performance of the proposed sequencer compared to state-of-the-art, and ii) interpretability of the cancer predictions made by the proposed sequencer through the generated critical response maps. 

\subsection{Experimental Setup}
\label{experimentalsetup}

\begin{table}[t]
\centering
\caption{Number of samples for corresponding malignancy scores, for three different datasets. The datasets were created based on how the radiologist malignancy rating 3 was treated i.e., Ignored (I), treated as benign (B), or treated as malignant (M).}
 \renewcommand{\arraystretch}{2}
\setlength{\tabcolsep}{2pt}
\begin{tabular}[C]{ C{1.4cm}| C{1.4cm}|C{1.4cm}|C{1.4cm}|C{1.4cm}}
\toprule[0.3mm]
Malignancy score & Size & I & B & M\\\hline 
1 & 1376 & \multirow{2}{*}{3981} & \multirow{3}{*}{9638} & \multirow{2}{*}{3981}\\ \cline {1-2}
2 & 2605 & & & \\ \cline{1-3} \cline{5-5}
3 & 5657 & Ignored & & \multirow{3}{*}{10452}\\ \cline{1-4}
4 & 3192 & \multirow{2}{*}{4795} & \multirow{2}{*}{4795} & \\ \cline{1-2}
5 & 1603 & & & \\
\toprule[0.3mm]
\end{tabular}
\label{malignancy_distribution}
\end{table}

The setup of the experiments in this study can be described as follows. The cropped lung nodule images with their corresponding malignancy scores are obtained from the LIDC-IDRI dataset as explained in section~\ref{dataset1}. The distribution of the malignancy scores in the dataset is described in Table~\ref{malignancy_distribution}. Malignancy scores $1$ and $2$ are considered as benign, whereas scores $4$ and $5$ are considered as Malignant. The malignancy score $3$ can be either considered as benign, malignant or ignored depending as done by previous studies in this field. In this paper, we have created three different datasets where the score $3$ is considered malignant (dataset: `M'), benign (dataset: `B'), and ignored (Dataset: `I'). The final dataset distribution is shown in Table~\ref{malignancy_distribution}. Each dataset was further divided into $80\%$ training data, $10\%$ for validation and $10\%$ testing data. The pre-processing and dataset distribution is similar to Xie et. al.~\cite{xie2018fusing}.

\begin{figure}[h]
\includegraphics[scale=0.75]{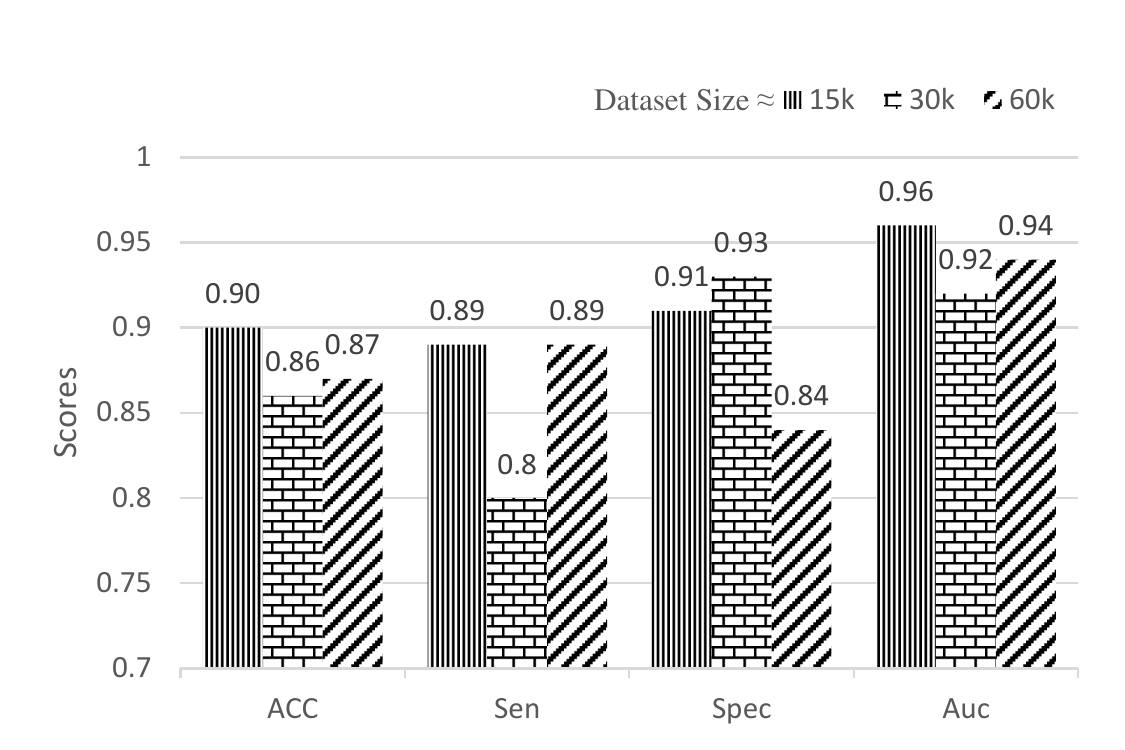}
\centering
\caption{Accuracy, sensitivity, and specificity for the ``ignored'' (I) dataset  for three different training sample sizes. It can be observed that the $\approx$15k size performed the best for 3 out of 4 metrics, including AUC; hence it was chosen as the default training size for the three different dataset categories: ignored (I), benign (B), and malignant (M).}
\label{diff_datasize_results}
\end{figure}

Table~\ref{dataset_aug_comparison} shows the distribution of training data for the three datasets. We can observe that dataset `B' and `M' are not evenly distributed. To mitigate this imbalance, data augmentation was performed on each of the datasets to balance the number of examples associated with each class.  Furthermore, the data augmentation performed also acts to enhance the variability and generalizability of the radiomic sequencer. In particular, random horizontal shifts, vertical shifts, and rotations were applied along with vertical and horizontal flips to construct the augmented training dataset. Since the size of the training data has a huge impact on the performance of the proposed radiomic sequencer, three different augmented datasets with varying size were created for each of the `M', `B' and `I' datasets, as shown in Table~\ref{dataset_aug_comparison}. The performance of dataset 'I' under different levels of data augmentation is shown in Fig~\ref{diff_datasize_results}. We can observe that the $\approx15$k size yielded the best performance. Going forward, we have finalized the dataset size to be $\approx15$k for further hyper-parameter tuning and validations for the `M', `B' and `I' datasets. 

Different data normalization techniques such as standard deviation, Min-Max, and ZCA whitening~\cite{krizhevsky2009learning} were also applied to the lung nodule images. It was found that Min-Max normalization yielded the best results. After optimizing the hyper-parameter using the validation set, batch normalization was implemented with momentum$= 0.99$ and dropout layer was used with rate$=0.25$.  The Adam optimizer was used with learning rate $= 1e^-5$ and batch size as $128$. The proposed radiomic sequencer was learnt for approximately $200$ epochs and evaluated against the test dataset. The final results were reported by averaging over 10-fold cross validation for all three datasets. 

\subsection{Cancer prediction performance}
\begin{figure}[h!]
Res\includegraphics[scale=0.5]{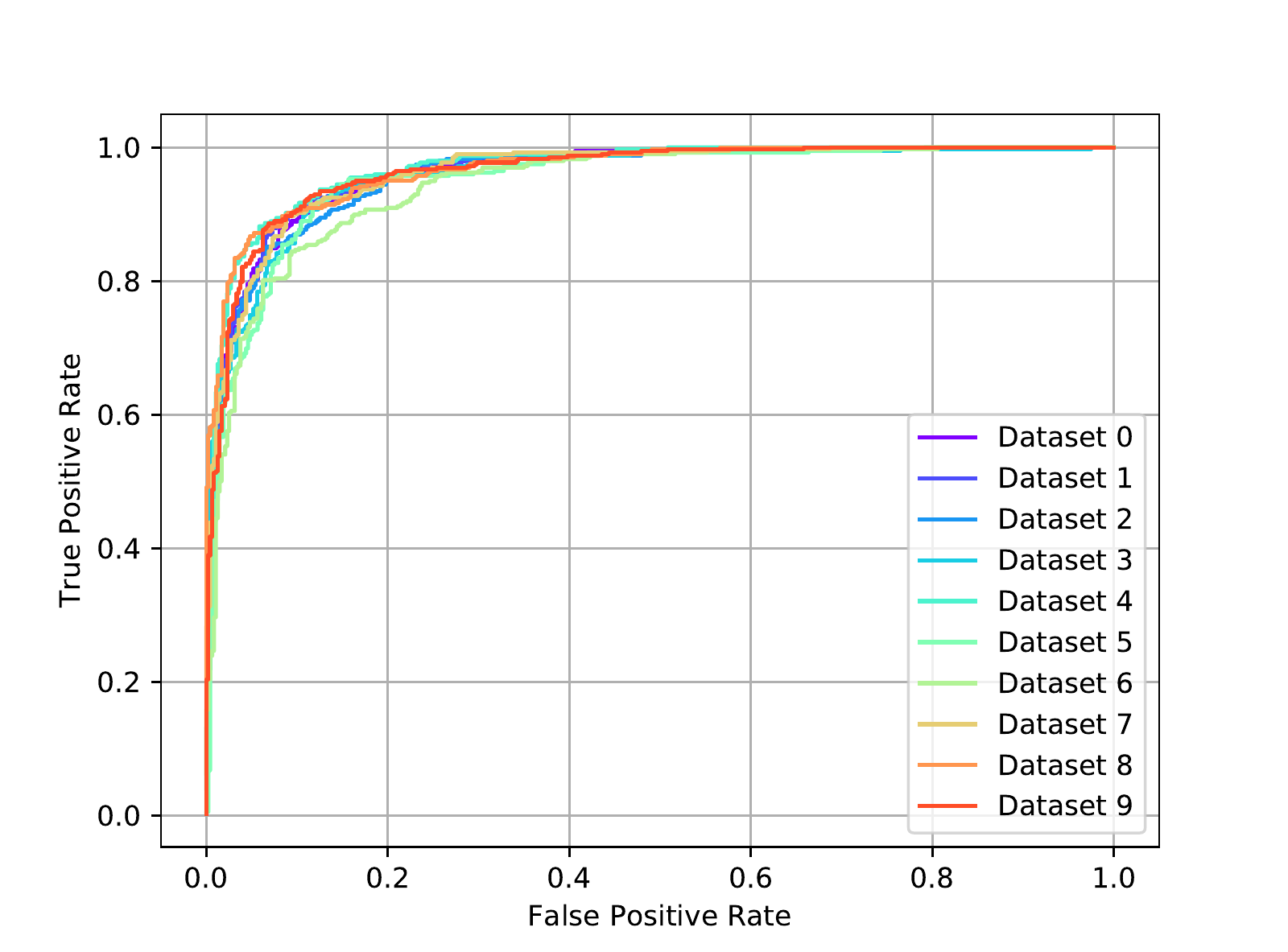}
\centering
\caption{Receiver operating curve (ROC) for the ignored (I) dataset for 10 different cross validation runs.}
\label{AucI10}
\end{figure}

\begin{figure}[h!]
\includegraphics[scale=0.5]{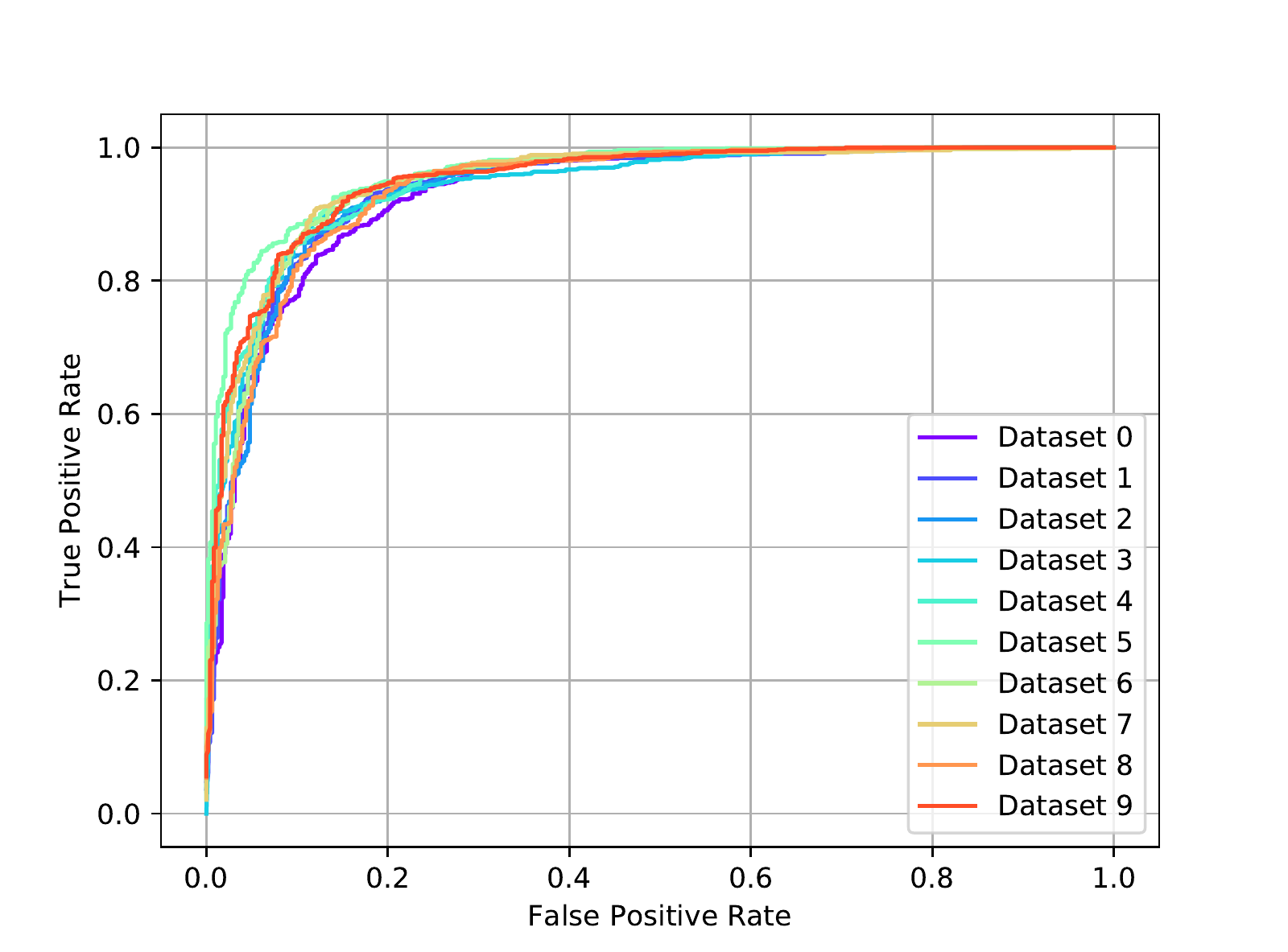}
\centering
\caption{Receiver operating curve (ROC) for the benign (B) dataset for 10 different cross validation runs.}
\label{AucB10}
\end{figure}

\begin{figure}[h!]
\includegraphics[scale=0.5]{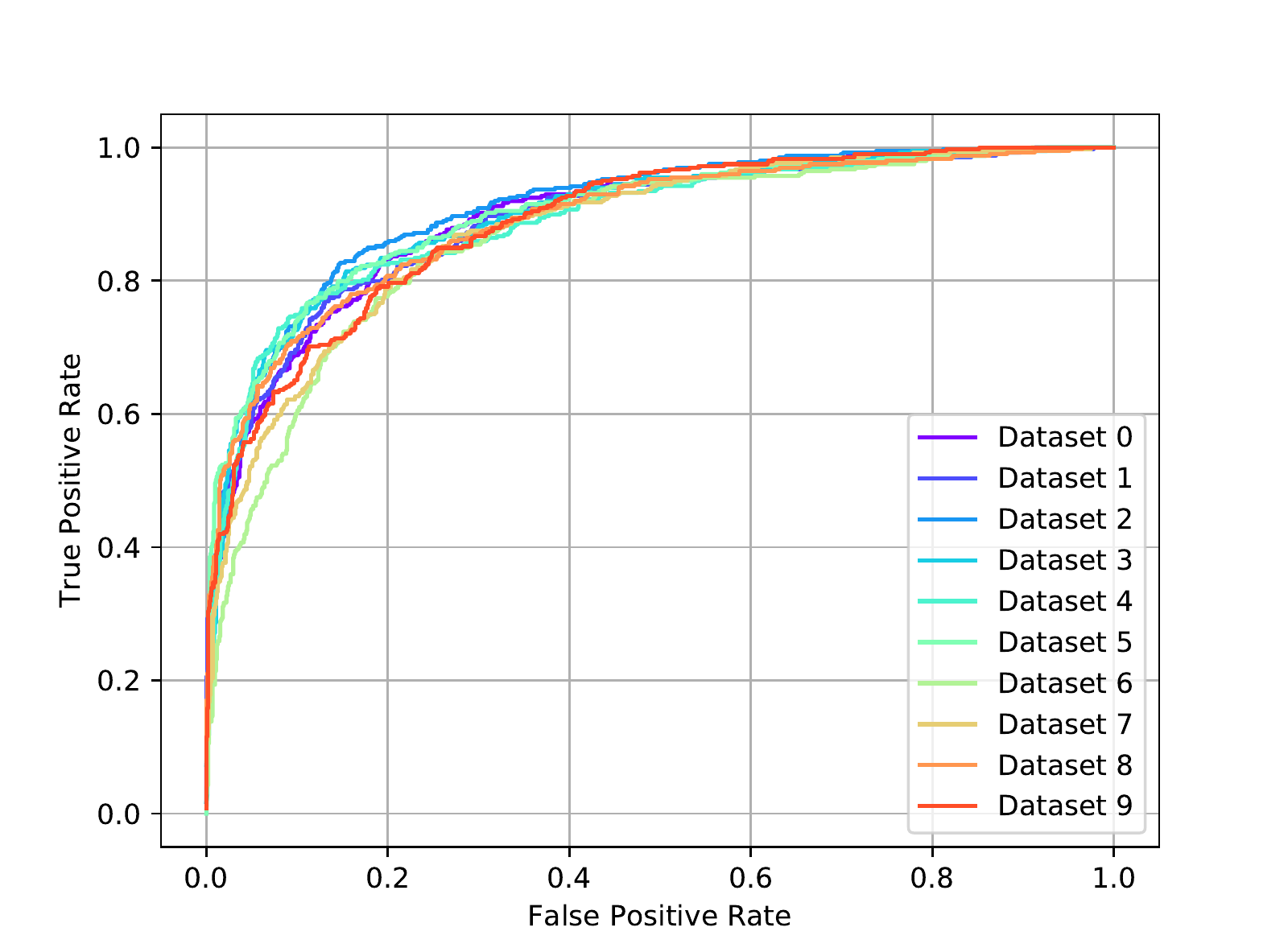}
\centering
\caption{Receiver operating curve (ROC) for the malignant (M) dataset for 10 different cross validation runs.}
\label{AucM10}
\end{figure}

\begin{figure}[t]
\includegraphics[scale=0.5]{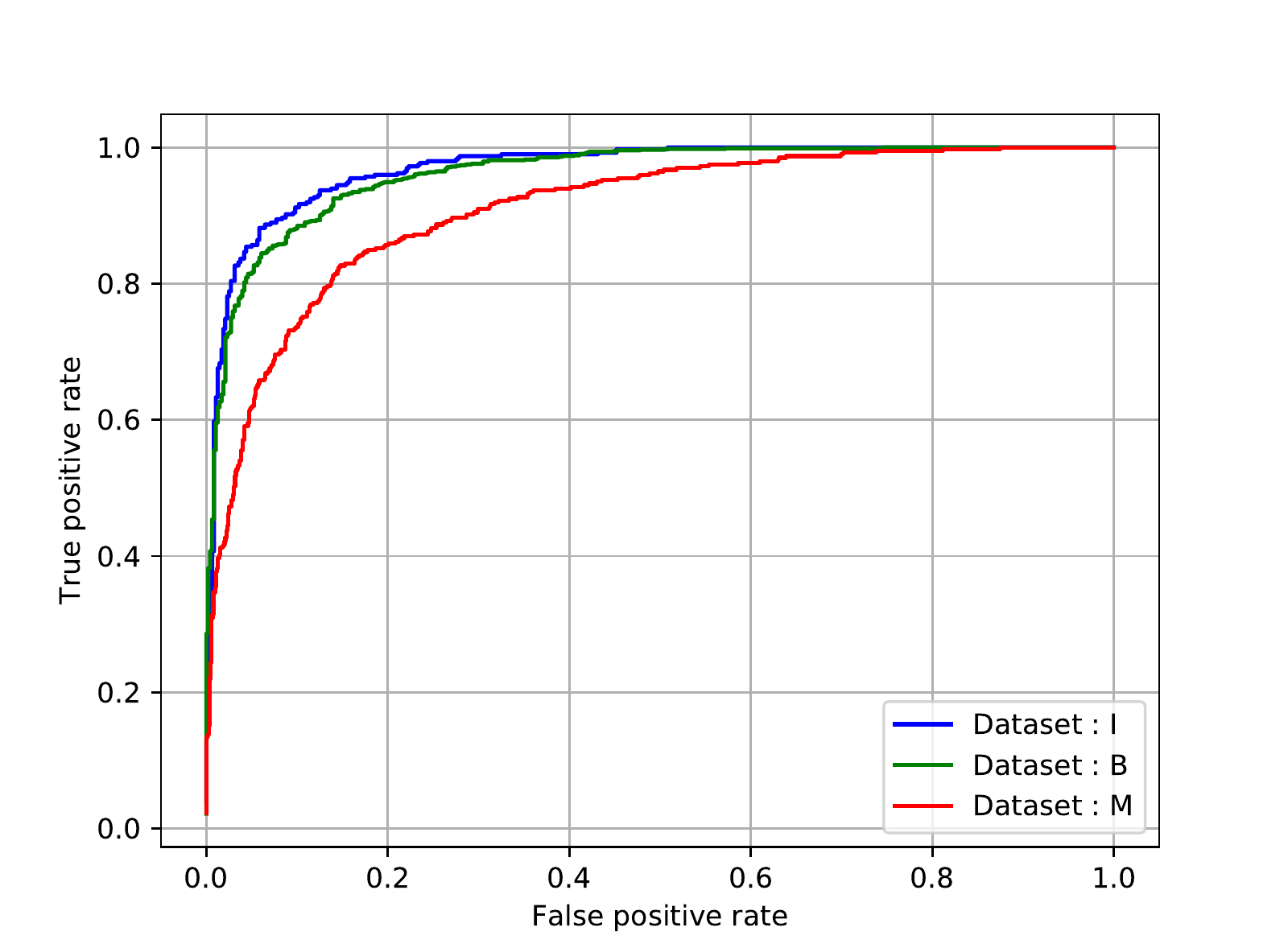}
\centering
\caption{Comparing the best ROC curves for the three different datasets: ignored (I), benign (B) and malignant (M).}
\label{bestAuc}
\end{figure}

To evaluate the cancer prediction performance of the proposed SISC radiomic sequencer, we computed the sensitivity, specificity, accuracy, and AUC of the proposed sequencer and compared it with five other state-of-the-art approaches. The average lung cancer prediction performance of the proposed radiomic sequencer for dataset `M', `B' and `I' are shown in Table.~\ref{M_dataset}~\&~\ref{B_dataset}~\&~\ref{I_dataset} respectively. The AUC curves of the 10 different cross validation runs for each dataset are shown in Figs.~\ref{AucI10},~\ref{AucB10}~\&~\ref{AucM10}. The best performing AUC curve from each dataset is shown in Fig.~\ref{bestAuc}.  From the results, we can observe that, for the dataset `I', by leveraging the proposed SISC radiomic sequencer, we were able to achieve comparable performance with the current state-of-the-art method proposed by Xie \etal~\cite{xie2018fusing}. For datasets `B' and `M', the proposed sequencer is able to outperform the accuracy results from Xie \etal~\cite{xie2018fusing}. The comparison of the existing and current state-of-the-art methods with the proposed SISC radiomic sequencer is given in Table.~\ref{M_dataset}~\&~\ref{B_dataset}~\&~\ref{I_dataset}. The comparison methods are based on the previous studies in this field, employing the same data pre--processing steps for fair comparison amongst the methods. Based on these experimental results, it can be observed that the proposed SISC radiomic sequencer can provide strong cancer prediction performance that exceeds state-of-the-art in all but one case, where in that case the performance is comparable to state-of-the-art.
 
\begin{table}[h!]
\centering
\caption{Dataset distribution for the three different dataset configuration obtained from the LIDC-IDRI dataset before and after data augmentation (as described in Section~\ref{experimentalsetup}).}
 \renewcommand{\arraystretch}{1.5}
\setlength{\tabcolsep}{1pt}
\begin{tabular}[C]{C{1.3cm} C{1.3cm} C{1.3cm} C{1.3cm} C{1.3cm} C{1.3cm}}  \toprule[0.3mm]
Dataset & Grade & Before Data Aug & 15k & 30k & 60k\\\hline
\multirow{3}{*}{I} & Malignant &3836 & 9836 & 15836 & 30836\\
 & Benign & 3184 & 9184 & 15184 & 30184 \\
& Total & 7020 & 19020	& 31020	& 61020 \\\hline
\multirow{3}{*}{B} & Malignant & 3836 & 7672 & 19180 &	38360\\
& Benign & 7712 & 7712 & 21712 & 35712\\
& Total & 11548 & 15384 & 40892 & 74072 \\\hline
\multirow{3}{*}{M} & Malignant & 8361 & 8361 & 16361 & 33444\\
& Benign & 3187 & 6374 & 15935 & 28683 \\
& Total & 11548 & 14735 & 32296 & 62127\\\toprule[0.3mm]
\end{tabular}
\label{dataset_aug_comparison}
\end{table}

\begin{table}[t!]
\centering
\caption{Performance comparison between tested cancer prediction methods for the ignored (I) dataset. Best results are highlighted in \textbf{bold}.}
 \renewcommand{\arraystretch}{2}
\setlength{\tabcolsep}{2pt}
\begin{tabular}[C]{C{2.1cm} C{1.5cm} C{1.5cm} C{1.5cm}C{1.5cm}}
\toprule[0.3mm]
Method & Accuracy                           & Sensitivity & Specificity & AUC 	\\\hline
Han \etal~\cite{han2013texture}	            & 85.59	& 70.62	& \textbf{93.02} & 89.25	\\
Dhara \etal~\cite{dhara2016combination}     & 88.38	& 84.58	& 90.03	& 95.76	\\
Shen \etal~\cite{shen2015multi}             & 87.14	& 77.00	& 93.00	& 93.00 \\
Sun \etal~\cite{sun2017balance}             & - & - & - & 88.23$\pm$1.70 \\
Xie \etal~\cite{xie2018fusing}           & \textbf{89.53$\pm$0.09} &	84.19$\pm$0.09 & 92.02$\pm$0.01 & \textbf{96.65$\pm$0.01} \\
Ours (SISC)                                   & 89.36$\pm$1.20& \textbf{90.28$\pm$2.00} &	88.25$\pm$2.00 & 96.01$\pm$0.70\\

\toprule[0.3mm]
\end{tabular}

\label{I_dataset}
\end{table}

\begin{table}[h!]
\centering
\caption{Performance comparison between tested cancer prediction methods for the benign (B) dataset. Best results are highlighted in \textbf{bold}.}
 \renewcommand{\arraystretch}{2}
\setlength{\tabcolsep}{2pt}
\begin{tabular}[C]{C{2.1cm} C{1.5cm} C{1.5cm} C{1.5cm}C{1.5cm}}
\toprule[0.3mm]
Method & Accuracy                           & Sensitivity & Specificity & AUC 	\\ \hline
Han \etal~\cite{han2013texture}	            & 87.36	&73.75&	93.37&	93.79	\\
Dhara \etal~\cite{dhara2016combination}     & 87.69	&80.00&	89.30&	94.44	\\
Xie \etal~\cite{xie2018fusing}           & 87.74$\pm$0.03&	\textbf{81.11$\pm$0.85}&	89.67$\pm$0.09&	\textbf{94.45$\pm$0.01} \\
Ours (SISC)  & \textbf{88.57$\pm$1.70} &	78.32$\pm$8.12 & \textbf{93.66$\pm$2.06} &	94.34$\pm$0.08\\
\toprule[0.3mm]
\end{tabular}
\label{B_dataset}
\end{table}

\begin{table}[h!]
\centering
\caption{Performance comparison between tested cancer prediction methods for the malignant (M) dataset. Best results are highlighted in \textbf{bold}.}
 \renewcommand{\arraystretch}{2}
\setlength{\tabcolsep}{2pt}
\begin{tabular}[C]{C{2.1cm} C{1.5cm} C{1.5cm} C{1.5cm}C{1.5cm}}
\toprule[0.3mm]
Method & Accuracy    & Sensitivity & Specificity & AUC 	\\\hline
Kumar \etal~\cite{kumar2015lung}     & 75.01 & 83.35 & - & - \\
Han \etal~\cite{han2013texture}	 & 70.97 & 53.61 & 89.41 & 76.26	\\
Dhara \etal~\cite{dhara2016combination}  & 71.17 & 53.47 & \textbf{89.74} & 79.74	\\
Sharma \etal~\cite{sharma2018early}        & 84.13 & \textbf{91.69} & 73.16 & - \\
Xie \etal~\cite{xie2018fusing} & 71.93$\pm$0.04 & 59.22$\pm$0.04 & 84.85$\pm$0.10& 81.24$\pm$0.01\\
Ours (SISC)			& \textbf{84.17$\pm$1.50}  & 90.71$\pm$4.01  & 67.00$\pm$8.64 & \textbf{89.06$\pm$1.20}\\
\toprule[0.3mm]
\end{tabular}
\label{M_dataset}
\end{table}

\subsection{Interpretability}

Here, we will investigate the efficacy of the proposed SISC radiomic sequencer in terms of interpretability of the lung cancer predictions made.  Fig.~\ref{Deconv1} shows example critical response maps generated in an end-to-end manner for several example malignant nodule images that were correctly predicted to be malignant. From the figure, it can be observed that the proposed SISC radiomic sequencer is able to successfully identify the nodule regions in the given CT slices without being explicitly directed to do so. As shown in Fig.~\ref{Deconv1}, the highlighted region's contour closely matches the contours given by the radiologists, and in some cases provide improved contour localization than that provided by the radiologists. The proposed SISC radiomic sequencer is able to successfully highlight a wide range of nodules with different shapes and sizes. We can also infer the discriminative nature of the proposed sequencer by observing that the highlighted regions in the critical response maps that contribute highly to a malignancy prediction. This helps to gain better insight in the rationale behind the malignancy prediction. Similar observations can also be observed for benign nodules as shown in Fig.\ref{Deconv0}. 

Due to the end-to-end interpretable nature of the proposed SISC radiomic sequencer, the critical response maps produced through the entire stack of interpretable sequencing cells can potentially help improve the confidence of radiologists working with the CAD system. Furthermore, the critical response maps can also assist radiologists to more consistently and rapidly spot abnormal nodules within the large volume of a CT scan, as well as understand the nature and characteristics most linked to malignancy. 

\section{Conclusion}
\label{conclusion}

In this paper, we introduce a novel end-to-end interpretable discovery radiomics-driven lung cancer prediction framework. This framework is enabled by the proposed radiomic sequencer: a deep stacked interpretable sequencing cell (SISC) architecture comprised of interpretable sequencing cells. Experimental results show that the proposed SISC radiomic sequencer is able to not only achieve state-of-the-art results in lung cancer prediction, but also offers prediction interpretability in the form of critical response maps generated through the stack of interpretable sequencing cells which highlights the critical regions used by the sequencer for making predictions. The critical response maps are useful for not only validating the predictions of the proposed SISC radiomic sequencer, but also provide improved radiologist-machine collaboration for improved diagnosis.


%

\section*{Acknowledgment}

The authors would like to thank NSERC, Canada Research Chairs program, and NVidia CEO Jen-Hsun Huang for providing the Nvidia Titan V CEO edition graphics processing unit used to run experiments in this study.

\ifCLASSOPTIONcaptionsoff
  \newpage
\fi

%




\bibliographystyle{unsrt}
\bibliography{sample}

\end{document}